%% file: article_EllipsoidalObstacle.tex
\documentclass[preprint, 3p]{elsarticle}
\journal{Mechatronics}

\usepackage{graphicx}       
\usepackage{amsmath,amssymb}
\usepackage{mathtools} 
\usepackage{bm}
\usepackage{xcolor}
\usepackage{natbib}
\newcommand*{\tran}{^{\mkern-1.5mu\mathsf{T}}} 

\usepackage{pgfplots}  
\pgfplotsset{compat=newest}
\usepackage{tikz, listofitems}

\usetikzlibrary{calc, shapes, backgrounds, bending, decorations.pathmorphing, positioning, decorations.pathreplacing, arrows.meta}
\usepgflibrary{arrows}
\usepackage{enumitem} 
\usepackage{units}

\begin{document}
\include{figures/header_plots}

\begin{frontmatter}
	\title{Efficient Avoidance of Ellipsoidal Obstacles with Model Predictive Control for Mobile Robots and Vehicles} 
	
	\author[First]{Mario Rosenfelder} 
	\author[First]{Hendrik Carius} 
    \author[Second]{Markus Herrmann-Wicklmayr}
	\author[First]{Peter Eberhard}
    \author[Second]{Kathrin Fla{\ss}kamp}
	\author[Third]{Henrik Ebel}
	
	\address[First]{Institute of Engineering and Computational Mechanics~(ITM), University of Stuttgart, Germany,\
	(e-mail: \lbrack mario.rosenfelder, peter.eberhard\rbrack @itm.uni-stuttgart.de; st171012@stud.uni-stuttgart.de)}
	\address[Second]{Systems Modeling and Simulation, Systems Engineering, Saarland University, Germany, (e-mail: [markus.herrmannwicklmayr, kathrin.flasskamp]@uni-saarland.de)}
	\address[Third]{Department of Mechanical Engineering, LUT University, Lappeenranta, Finland,\
	(e-mail: henrik.ebel@lut.fi)}
    
	\begin{abstract}
    In real-world applications of mobile robots, collision avoidance is of critical importance.
    Typically, global motion planning in constrained environments is addressed through high-level control schemes.
    However, additionally integrating local collision avoidance into robot motion control offers significant advantages.
    For instance, it reduces the reliance on heuristics and conservatism that can arise from a two-stage approach separating local collision avoidance and control. 
    Moreover, using model predictive control (MPC), a robot's full potential can be harnessed by considering jointly local collision avoidance, the robot's dynamics, and actuation constraints. 
    In this context, the present paper focuses on obstacle avoidance for wheeled mobile robots, where both the robot's and obstacles' occupied volumes are modeled as ellipsoids.
    To this end, a computationally efficient overlap test, that works for arbitrary ellipsoids, is conducted and novelly integrated into the MPC framework.
    We propose a particularly efficient implementation tailored to robots moving in the plane. 
    The functionality of the proposed obstacle-avoiding MPC is demonstrated for two exemplary types of kinematics by means of simulations. 
    A hardware experiment using a real-world wheeled mobile robot shows transferability to reality and real-time applicability. 
    The general computational approach to ellipsoidal obstacle avoidance can also be applied to other robotic systems and vehicles as well as three-dimensional scenarios. 
	\end{abstract}

	\end{frontmatter}
\section{Introduction}\label{sec:Intro}
In real-world scenarios, mobile robots and vehicles have to navigate without colliding with obstacles by all means.
Static, large-scale obstacles have usually been considered at a high-level planning stage already, based on a predefined or SLAM-identified map of the environment; consider, for instance, the navigation  through corridors of a building. 
In contrast, individual obstacles of smaller scale, moving or of non-permanent character, are typically dealt with using local, reactive approaches, including heuristics-
based techniques such as artificial potential field methods~\citep{khatib1986real}. 
Alternatively, there exists rigorous control theoretical approaches such as control barrier functions~\citep{Ames2019}. 
In practice, both can yield to the same type of system behavior for obstacle avoidance and they are often used to alter control inputs coming from controllers that, themselves, do not account for obstacles and were not designed with obstacles in mind. 

However, integrating motion control with (local) obstacle avoidance can have several benefits. 
Not only does it simplify the control structure, eases tuning, and prevents a waste of efficiency or performance arising from an artificial separation of the tasks, but also does it helps to resolve particularly critical situations, e.g., when a robot of non-circular footprint needs to rotate to squeeze through a narrow passageway between obstacles, by intrinsically coupling dynamic control and obstacle avoidance. 

In formal motion planning language, obstacles are represented as constraints on the variables defining the working space of (mobile) robots. 
 Despite the aforementioned potential advantages, obstacle avoidance is usually not integrated into dynamic real-time closed-loop control because few control approaches can deal with constraints in a natural and efficient manner. 
In control settings, collisions must not only be detected or evaluated, as it is common in computer graphics or particle simulations, but also proactively avoided in real time by selecting an appropriate (optimal) control input.
Indeed, probably the only real-time capable control approach which is able to straight-forwardly  deal with constraints without any conservativity is model predictive control (MPC). 
However, constraints as they arise from obstacles are typically expensive to evaluate and naturally non-convex. 
Thus, while it is simple to set-up an MPC incorporating obstacles as constraints, presently, one typically 
cannot reliably solve the non-convex optimal control problem therein in real time.
For instance, in the recent work~\cite{Celestini24}, in a setting similar but not identical to the one considered in this work, a machine-learning scheme 
has to provide initial guesses and a terminal cost for an MPC controller.
The authors note that, in their hardware experiments, their ML-based version was the only one to ``not require loosened time constraints'', i.e., lowered control frequency, unlike conventional MPC implementations did. 

In contrast, in this paper, we propose to integrate collision avoidance directly into MPC problems without having to add any machine-learning component.  We obtain a real-time capable performance by considering robots and obstacles of ellipsoidal shape, which provide a collision constraint in terms of intersecting ellipsoids.
We implement this constraint efficiently specifically for the planar case for direct integration into MPC problems, further speeding up solution compared to a general implementation. 
The planar case is of high practical relevance, e.g., in service robotics, and thus warrants a tailored implementation. 
To the authors' knowledge, this is the first time that such an MPC controller was proposed and successfully tested in real-time with robotic hardware to successfully avoid collisions between an ellipsoidal mobile robot and ellipsoidal obstacles where the robot can also freely rotate, without any of the involved shapes having to be circular or axis aligned. 
Classical other benefits of using MPC for robot control are carried over from obstacle-free applications, e.g., the explicit consideration of input bounds for optimal performance also at the edge of the robots' capabilities. 

The paper is organized as follows. 
In Sec.~\ref{sec:ellipsoid} we recapitulate an efficient way to test for overlaps of ellipsoids from~\cite{gilitschenski2012, gilitschenski2014} and tailor it for our use case with MPC. 
In Sec.~\ref{sec:mpc}, we describe how the approach can be integrated into the constraints of an optimal control problem (OCP) for MPC. 
Section~\ref{sec:results} provides results in the form of a simulative study and real-time experiments with robot hardware, whereas Sec.~\ref{sec:conclusions} concludes the paper. 

\section{Efficient Overlap Testing of Ellipsoids}\label{sec:ellipsoid}
Our aim is to realize obstacle avoidance by adding constraints to the OCP formulation used in a predictive control setting. 
Basically, we need constraints for each obstacle that enforce the distance between robot and obstacle being larger than zero or than a positive safety margin at each time step over the controller's prediction horizon.
In this paper, we assume that collision avoidance at the chosen sampling instances is sufficient, i.e.\ we assume that inter-sample deviations are taken care of by a safety margin and sufficiently small sampling rates. 
Importantly, the distance-like quantity used in the constraints does not need to be the actual (Euclidean) distance, it is sufficient to choose a metric that behaves similar within the relevant cases. For instance, the distance might be put in direct correlation with the overlap, i.e., it may be zero for zero overlap, positive for an overlap of positive volume, negative for no overlap and ideally (strictly) monotonic and smooth to help the numerical solution algorithm. 
As the overlap-like or distance-like quantity is evaluated many times in the iterative OCP solution process, its evaluation must be as time efficient as possible. 
However, calculating the actual distance (or overlap) between two arbitrary ellipsoids is non-trivial. 
While it can be stated easily as an optimization problem itself, nesting one optimization problem in another will not be efficient. 
In another approach, one may calculate the Minkowski sum of the two ellipsoids and then equivalently evaluate the distance of a point to the sum. 
However, whereas it is easy to calculate the distance of a point to an ellipsoid, the Minkowski sum of two ellipsoids is generally not an ellipsoid~\citep{Yamada2024}, and an approach to obtain the sum is, again, the solution of a non-trivial optimization problem~\citep{Halder2018}. 
Therefore, we build upon~\cite{gilitschenski2012, gilitschenski2014}, who propose a method to efficiently test for the overlap of two (arbitrary-dimensional) ellipsoids for usage in data fusion and filtering. 
Therein, it is only detected whether there is an overlap or not. 
In contrast, we are interested in an overlap-like metric following from this approach quantifying how much the ellipsoids are overlapping or not, see also~\citep{Ros2002}. 
Although for our immediate use case, only two-dimensional ellipsoids are of interest, we first consider the arbitrary-dimensional case. 

\begin{figure}
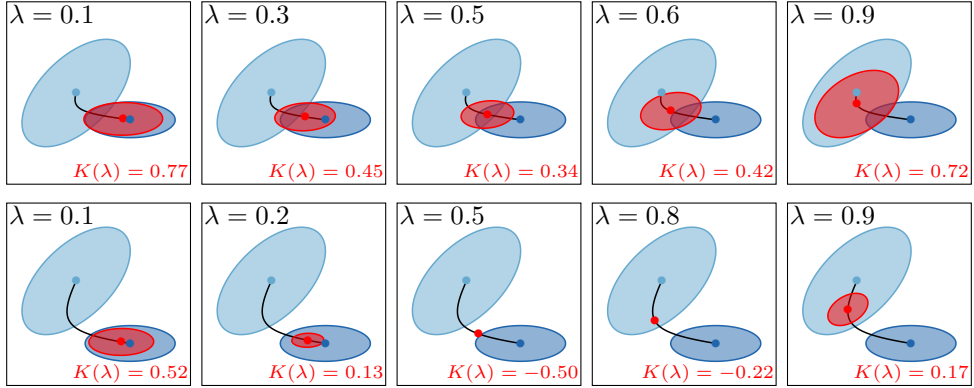
%
	\centering%
	\include{figures/figure_ellipsoids}%
	\caption{Resulting ellipsoid $\mathcal{E}_\lambda$ (red) following from the original ellipsoids $\mathcal{E}_1$ and $\mathcal{E}_2$ (blue) for different values of $\lambda\in(0,1)$ in case of collision (top row) or no collision (bottom row).}\label{fig:ellipsoid_lambda}
\end{figure}

Consider an arbitrary ellipsoid of dimension $n$, which can be described by means of
\begin{align}\label{eq:ellipsoid}
	\mathcal{E}&(\bm{M},\bm{p}) \coloneqq \left\lbrace \bm{x}\in\mathbb{R}^n: \, (\bm{x}-\bm{p})\tran \bm{M} (\bm{x}-\bm{p}) \leq 1 \right\rbrace, 
\end{align} 
where $\bm{M}\in\mathbb{R}^{n\times n}$ is the symmetric positive definite ellipsoid matrix and $\bm{p}\in\mathbb{R}^n$ the location vector of the ellipsoid's center.
In order to avoid the collision of a robot with an obstacle, we want to ensure that the intersection of two ellipsoids $\mathcal{E}_1 (\bm{A},\bm{v})$, $\mathcal{E}_2 (\bm{B},\bm{w})$ is empty.
To this end, a new ellipsoidal set is defined~\citep{gilitschenski2012,gilitschenski2014} with respect to a parameter $\lambda\in(0,1)$ reading
\begin{subequations}
\begin{align}\label{eq:ellipsoid_lambda}
	\mathcal{E}_\lambda (\mathcal{E}_1, \mathcal{E}_2) &\coloneqq \lbrace \bm{x}\in\mathbb{R}^n: \, (\bm{x}-\bm{m}_\lambda)\tran \bm{E}_\lambda (\bm{x}-\bm{m}_\lambda) \leq K(\lambda) \rbrace,  \\
	\bm{E}_\lambda &\coloneqq \lambda \bm{A} +(1-\lambda)\bm{B}, \label{eq:ellipsoid_lambda_comp}\\
	\bm{m}_\lambda &\coloneqq \bm{E}_\lambda^{-1} \left( \lambda \bm{A} \bm{v} + (1-\lambda)\bm{B}\bm{w} \right),  \\
	K(\lambda) &\coloneqq 1-\lambda\bm{v}\tran\bm{A}\bm{v} - (1-\lambda)\bm{w}\tran\bm{B}\bm{w} + \bm{m}_\lambda\tran \bm{E}_\lambda \bm{m}_\lambda. 
\end{align}
\end{subequations}
As illustrated in Fig.~\ref{fig:ellipsoid_lambda} for $n=2$, the set~\eqref{eq:ellipsoid_lambda} transfers for increasing $\lambda$ ellipsoid~$\mathcal{E}_2$ into ellipsoid~$\mathcal{E}_1$ while always containing the intersection of both such that it holds that
$\left( \mathcal{E}_1 \cap \mathcal{E}_2 \right) \subseteq \mathcal{E}_\lambda \subseteq \left( \mathcal{E}_1 \cup \mathcal{E}_2 \right)$,
where here and in the following, the arguments of the ellipsoids are mostly omitted for reasons of readability.
By definition, for $\lambda=0$, it holds that $\mathcal{E}_\lambda=\mathcal{E}_2$ and analogously $\mathcal{E}_\lambda=\mathcal{E}_1$ for $\lambda=1$.
Overall, the expression describing the ellipsoid $\mathcal{E}_\lambda$ for some fixed $\lambda\in(0,1)$ is composed as follows. 
The ellipsoidal matrix~$\bm{E}_\lambda$ follows as the convex combination of the ellipsoid matrices of the two original ellipsoids. 
While the location vector~$\bm{m}_\lambda$ is easy to interpret, see also its course w.r.t. $\lambda$ as illustrated by the black line in Fig.~\ref{fig:ellipsoid_lambda}, it is in general computationally expensive since it requires the inversion of an $n$-dimensional matrix.
Thirdly, $K(\lambda)\in\mathbb{R}$ defines the size of the resulting ellipsoid. 
Crucially, $K(\lambda)\in\mathbb{R}$ can also take non-positive values for some $\lambda$, meaning that, then, set~\eqref{eq:ellipsoid_lambda} is empty, cf. Fig.~\ref{fig:ellipsoid_lambda} (bottom) for $\lambda=\{0.5, \, 0.8 \}$ and the digital supplementary material, which contains an illustrative animation. 
Thus, there is no collision if there exists a $\lambda\in(0,1)$ such that $K(\lambda)<0$.

Hence, to check for the overlap of the two ellipsoids, we can determine the minimum of $K(\lambda)$ on the interval $\lambda\in(0,1)$, where we denote the minimizer by $\lambda^\star$.
If one is only interested in a binary collision check, considering the sign of $K(\lambda^\star)$ is sufficient, but we will consider its explicit value for MPC.
\begin{figure}
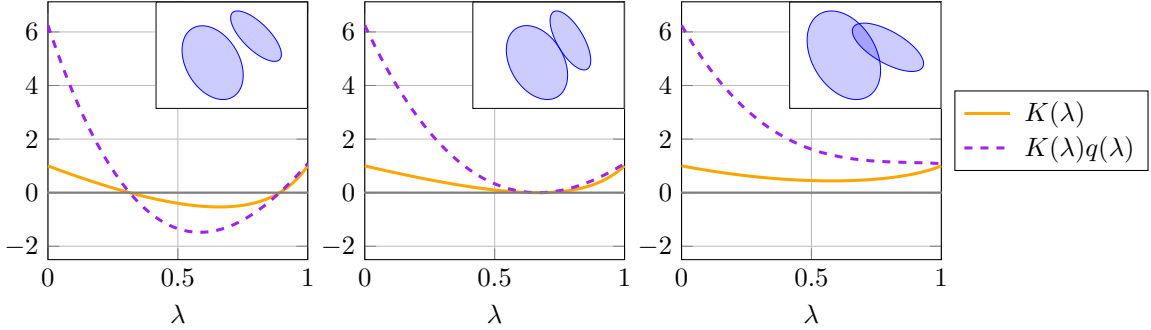

	\centering
    \include{figures/figure_K}%
	\caption{Crucial metrics $K(\lambda)$ and $K(\lambda)q(\lambda)$ for no contact or overlap (left), contact (middle), and overlap (right).}%
	\label{fig:Kq_lambda}%
\end{figure}

The minimum $\lambda^\star$ of $K(\lambda)$ is hard to directly calculate explicitly as $K(\lambda)$ is a rational fraction. 
Thus, we will use a quantity of similar character as $K(\lambda)$ to assess overlaps. 
To this end, we define $\bm{E}_\lambda^{-1}=\frac{\text{adj}(\bm{E}_\lambda)}{\text{det}(\bm{E}_\lambda)}\eqqcolon {\bm{R}(\lambda)}/{q(\lambda)}$.
Substituting $\bm{m}_\lambda$ using the previous definition into $K(\lambda)$ then yields
\begin{align}\label{eq:Kq}
	K(\lambda) q(\lambda) &= q(\lambda) \left( 1-\lambda \bm{v}\tran\bm{A}\bm{v} - (1-\lambda)\bm{w}\tran\bm{B}\bm{w} \right) + \bm{u}(\lambda)\tran \bm{R}(\lambda) \bm{u}(\lambda), 
\end{align}
where $\bm{u}(\lambda)\coloneqq  \lambda \bm{A}\bm{v}+(1-\lambda)\bm{B}\bm{w}$.
For $n=2$, the determinant $q(\lambda)$ and adjoint $\bm{R}(\lambda)$ can be explicitly obtained as a quadratic polynomial and a $\mathbb{R}^{2\times2}$ matrix affine in~$\lambda$, respectively. Thus, we can write
\begin{align}\label{eq:Kq_cubic}
	K(\lambda) q(\lambda) &= h_3 \lambda^3 + h_2 \lambda^2 + h_1 \lambda + h_0 \eqqcolon f(\lambda,(\bm{A},\bm{v}), (\bm{B},\bm{w}) ) = f(\lambda,\mathcal{E}_1, \mathcal{E}_2), 
\end{align} 
where $h_i\in\mathbb{R}$, $i\in\{0,\dots,3\}$.
Since the determinant~$q(\lambda)$ is greater than zero for all $\lambda$ due to the matrix~$\bm{E}_\lambda$ being positive definite as a convex combination of positive definite matrices, $K(\lambda)q(\lambda)$ attains the same roots as $K(\lambda)$ such that the sign of $K(\lambda^\star)q(\lambda^\star)$ and $K(\lambda^\star)$ are equal, meaning we can and will use the value and sign of the simpler expression $K(\lambda^\star)q(\lambda^\star)$ in a similar manner as the one of $K(\lambda^\star)$ to assess overlaps.

If existing, the minimum of the right-hand side of Eq.~\eqref{eq:Kq_cubic}, can be determined explicitly as
\begin{align}\label{eq:lambda_star}
	\lambda^\star = \frac{-h_2 + \sqrt{h_2^2 -3h_3 h_1} }{3h_3}.
\end{align}
Note that there exist two special cases that need to be checked before evaluating $\lambda^\star$.
Firstly, if the function is quadratic only ($h_3=0$), then $\lambda^\star = -{h_1}/{2h_2}$ and, secondly, if $(h_2^2 - 3h_3 h_1)<0$, then the cubic function does not attain a minimum on the interval $\lambda\in(0,1)$.
In this case, the minimum on the closed interval of interest is attained at one of the boundaries, i.e., $\lambda^\star \in \{0,1\}$, and takes a positive value, implying that the ellipsoids overlap. 
With $\lambda^\star$, we obtain the value of the overlap-like metric $K(\lambda^\star)q(\lambda^\star)$, which we abbreviate in the following as $\kappa^{\star}\coloneqq K(\lambda^\star) q(\lambda^\star)$.
To ensure collision avoidance, by construction, $\kappa^\star<0$ must hold, see also Fig.~\ref{fig:Kq_lambda}.

Alternatives to the explicit calculation of the minimum exist, independently of~$n$. 
These include a conservative approximation using the minimum value of a finite amount of sample points and employing numerical optimization to obtain the minimum.
These approaches are for example promising for the three-dimensional setup.
However, for $n=2$, the computationally cheapest option appears to be the previous explicit consideration.
Thus, this formulation is used subsequently.

\section{Obstacle-Avoidant Model Predictive Controller}
\label{sec:mpc}
We consider quite generic robot or vehicle kinematics of the form
\begin{align}\label{eq:kinematics_robot}
    \dot{\bm{x}} = \bm{G}(\bm{x}) \, \bm{u}
\end{align} where $\bm{x}\in\mathcal{X}\subset \mathbb{R}^{n_x}$ comprises the robot's state and $\bm{u}\in\mathcal{U}\subset\mathbb{R}^{n_u}$ its inputs.
The kinematics of all typical mobile robots and vehicles can be expressed in this form, including nonholomic ones, cf.~\cite{RoseEbel23}.
Moreover, we assume w.l.o.g.\ that the state~$\bm{x}$ comprises the robot's position in the plane within the first two components, i.e., $\bm{x}_{1:2} = \begin{bmatrix} x_1 & x_2\end{bmatrix}\tran$ and its orientation as the third coordinate, i.e., $x_3 = \theta$.

The kinematics~\eqref{eq:kinematics_robot} is then used to formulate the model predictive controller.
As mobile robots are mostly operated in a discrete-time fashion with piecewise constant control inputs over a sampling time $\delta_t\in\mathbb{R}_{>0}$, we consider the discrete-time case.
Using typical MPC notation in which $\bm{x}(k\,\vert\,t)$ and $\bm{u}(k\,\vert\,t)$ denote the predicted state and input trajectories of the discretized system planned at time $t$ and evaluated at time steps $k$, the optimal control problem follows as
\begin{subequations}\label{eq:OCP}
    \begin{alignat}{3}%
            &\hspace{0pt} \underset{\bm{u}(\cdot\,\vert\, t)}{\text{minimize}}
            &&\hspace{6pt}\!\sum_{k=t}^{t+H}\!\ell\left(\bm{x}(k\,\vert\, t),\bm{u}(k\,\vert\, t)\right)\label{eq:mpc_cost_generic}\\
            & \text{subject to}
            &&\hspace{6pt}{\bm{x}}(k+1\,\vert\, t)= \bm{f}^{\delta_t}_{\textnormal{d}}(\bm{x}(k\,\vert\, t), \bm{u}(k\,\vert\, t)), \label{eq:mpc_dynamics_generic}\\
            &&&\hspace{6pt} \bm{u}(k \,\vert\, t)\in\mathcal{U}, \label{eq:mpc_input_constraints}\\
            &&&\hspace{6pt} \bm{x}(k \,\vert\, t)\in\mathcal{X}\quad \forall k\in\mathcal{H},\label{eq:mpc_state_constraints} \\
            &&&\hspace{6pt}\bm{x}(t\,\vert\, t)=\bm{x}(t)\label{eq:mpc_measurement}
    \end{alignat}%
    \label{eq:mpc_optprob_generic}%
\end{subequations}%
with $\mathcal{H}\coloneqq\lbrace t,\dots, t+H\rbrace$, some continuous stage cost $\ell:\,\mathbb{R}^{n_x}\times\mathbb{R}^{n_u}\to\mathbb{R}_{\geq 0}$ summed over the prediction horizon of length $H\in\mathbb{N}$, and as prediction model~\eqref{eq:mpc_dynamics_generic} some (approximate) discretization of~\eqref{eq:kinematics_robot}.
Then, the implicit control strategy follows by solving the OCP~\eqref{eq:OCP} at each time instant $t\coloneqq j\delta_t$, $j\in\mathbb{N}_0$, applying the first sequence of the optimal input over the sampling interval, i.e., $\bm{u}(t)\coloneqq \bm{u}^\star(t)$ for $t\in[j\delta_t, (j+1)\delta_t)$, and then repeating this procedure. 
Mostly, we do not explicitly state the variables' time-dependency for brevity.

Focal point of this paper is to include obstacle avoidance through the OCP's state constraints~\eqref{eq:mpc_state_constraints} in an efficient manner. 
We add one state constraint per obstacle and discrete time step along the prediction horizon. 
Thus, subsequently, we describe how said constraint is furnished for a single obstacle and time step. 
To that purpose, let the volume occupied by the robot be described by some ellipsoid $\mathcal{E}(\bm{R}(\bm{x}_3),\bm{x}_{1:2})\eqqcolon\mathcal{E}_\textnormal{r}(\bm{x})$, cf. Eq.~\eqref{eq:ellipsoid}, where the planar ellipsoid's center is determined by the robot's position $\bm{x}_{1:2}$ and the ellipsoid matrix~$\bm{R}$ is a function of the robot orientation. 
Further, the obstacle's occupied volume is also described by a planar ellipsoid $\mathcal{E}_\textnormal{c}\coloneqq \mathcal{E}(\bm{D},\bm{e})$. 
In this work, the obstacles studied are static, but the method can be naturally extended to include (known) time dependencies. 
Both ellipsoids are assumed to not change their shapes over the course of time, and they can be outer approximations of robot and obstacle, respectively, for instance to take into account safety margins or accommodate more general shapes. 
Then, obstacle avoidance at the sampling instances between robot and studied obstacle is achieved by adding the constraint 
\begin{align}\label{eq:MPC_condition}
	f\left(\lambda^\star, \mathcal{E}_\textnormal{r}(\bm{x}(k\,\vert\,t)), \mathcal{E}_c \right) \overset{!}{<} 0,
\end{align} 
cf. Eq.~\eqref{eq:Kq_cubic}, for all time instants $k\in\mathcal{H}$ over the prediction horizon to the state constraint set~\eqref{eq:mpc_state_constraints}.
The functionality and computational efficiency of the proposed approach is shown in the following. 

\section{Results}
\def\indOm{om}
\def\indDD{d}
\label{sec:results}
Firstly, we consider an omnidirectional mobile robot such that its nominal kinematics can be described by a single-integrator $\dot{\bm{x}}=\bm{u}_{\textnormal{\indOm}}$ with the state $\bm{x}=\begin{bmatrix} x_1 & x_2 & \theta \end{bmatrix}\tran\in\mathcal{X}\subset \mathbb{R}^3$.
Due to the physical limitations of the employed electric motors, the macroscopic input $\bm{u}_{\textnormal{\indOm}}=\begin{bmatrix} v_{x_1} & v_{x_2} & \omega\end{bmatrix}\tran$, comprising the two translational velocities as well as the angular yaw velocity, is constrained so that the maximum absolute values of translational and angular velocities are $\unitfrac[0.2]{m}{s}$ and $\unitfrac[\nicefrac{\pi}{4}]{rad}{s}$, respectively.
In all following simulation and hardware experiments, the robot shall drive to its desired setpoint while evading obstacles in the direct line-of-sight of the initial pose and the desired setpoint, which, w.l.o.g., is assumed to be the origin. 
Hence, the conventional quadratic cost function $\ell(\bm{x},\bm{u}_{\textnormal{\indOm}})=\bm{x}\tran \bar{\bm{Q}} \bm{x} + \bm{u}_{\textnormal{\indOm}}\tran \bar{\bm{R}} \bm{u}_{\textnormal{\indOm}}$, $\bar{\bm{Q}}, \bar{\bm{R}}\succ 0$ is employed for the omnidirectional robot.
Secondly, we consider a differential-drive mobile robot described through
\begin{align} \label{eq:kinematics_dd}
    \dot{\bm{x}} = \begin{bmatrix}
        \dot{x}_1 \\ \dot{x}_2 \\ \dot{\theta}
    \end{bmatrix} = \begin{bmatrix}
        \cos \theta & 0 \\ \sin \theta & 0 \\ 0 & 1
    \end{bmatrix} \bm{u}_{\textnormal{\indDD}},
\end{align} where the input $\bm{u}_{\textnormal{\indDD}}=\begin{bmatrix}v & \theta \end{bmatrix}\tran\in\mathcal{U}_{\textnormal{\indDD}}\subset\mathbb{R}^2$, only consists of the translational and rotational velocities which are analogously constrained so that the maximum admissible absolute values are $\unitfrac[0.2]{m}{s}$ and $\unitfrac[\nicefrac{\pi}{4}]{rad}{s}$, respectively.
Deriving a stabilizing MPC for this nonholonomic robot is, even in the absence of obstacles, much more sophisticated but can be achieved by utilizing the tailored, mixed-exponents cost function $\ell(\bm{x},\bm{u}_{\textnormal{\indDD}})=q_1 x_1^4 + q_2 x_2^2 + q_3 \theta^4 + r_1 v^4 + r_2 \omega^4$, $q_i,\, r_i>0$, within OCP~\eqref{eq:OCP}, see~\cite{RoseEbel23,EbelRose23} for details.

In all simulation and hardware experiments, the sampling time is chosen to $\delta_t=\unit[0.2]{s}$ and the prediction horizon is set to $H=10$ steps resulting in a prediction horizon of $T=\unit[2]{s}$.
The resulting OCPs are formulated using CasADi~\citep{AndGill19} and solved with Ipopt~\citep{WachBieg06}, both through the Matlab interface.

\begin{figure*}
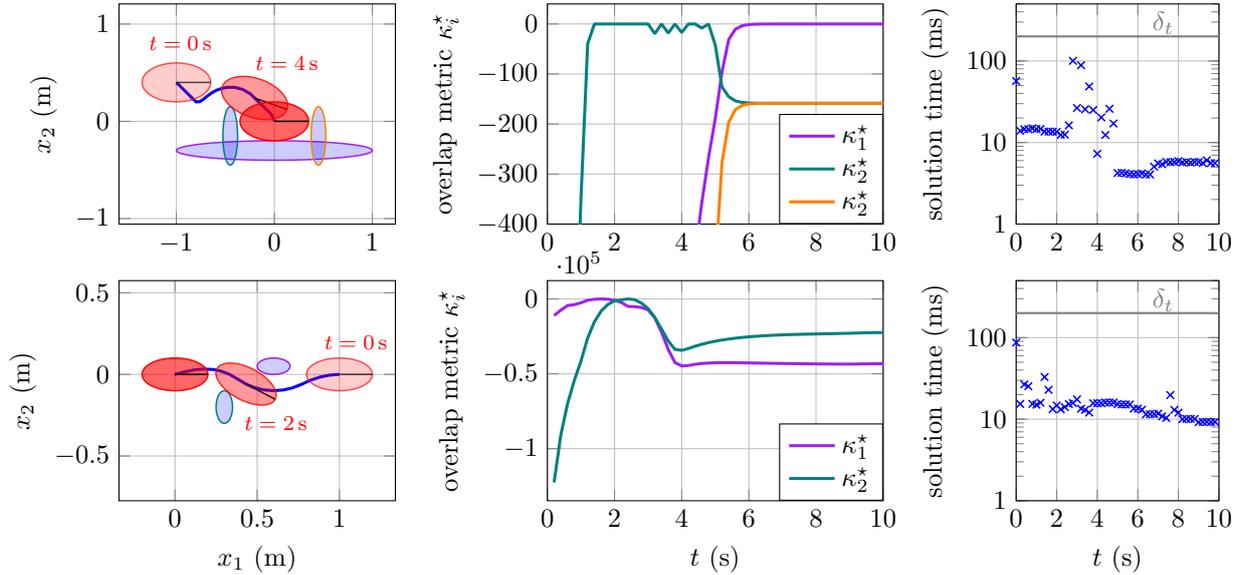

    \centering
    \include{figures/simulation_fig}
    \caption{Simulation results for the omnidirectional mobile robot (top plots) and the differential-drive mobile robot (bottom plots) containing a planar perspective including the obstacles (left plots), the crucial $K(\lambda^\star)q(\lambda^\star)$ metric (middle plots), and the solution times of the OCPs (right plots).}
    \label{fig:simulation}
\end{figure*}
\subsection{Simulation Results}
To illustrate the principal functionality of the proposed obstacle-avoidant MPC, we investigate a simulation scenario for an omnidirectional robot and for a differential-drive robot, both with their nominal dynamics. 
First, the omnidirectional robot starts at the initial pose $\bm{x}_0 = \begin{bmatrix} \unit[-1]{m} & \unit[0.4]{m} & \unit[0]{rad}\end{bmatrix}\tran$ and shall park at the origin.
Here, the robot's volume 
is an ellipsoid with semi-axis lengths of $\unit[0.35]{m}$ and $\unit[0.2]{m}$ yielding~$\bm{R}(x_3)$ such that its ellipsoid follows as $\mathcal{E}_\textnormal{r}(\bm{x})=\mathcal{E}(\bm{R}(x_3),\bm{x}_{1:2})$.
The three static obstacles present are described by means of $\mathcal{E}_{\textnormal{c},i}=\mathcal{E}(\bm{D}_i,\bm{e}_i)$, $i\in\{1,2,3\}$, as depicted in Fig.~\ref{fig:simulation}. 
As Fig.~\ref{fig:simulation} (top) indicates, the proposed controller successfully steers the robot to the origin while evading the obstacles.
In order to minimize the cost function~$\ell$ over the prediction horizon, the robot seeks to minimize its distance to the origin which includes that it rotates while passing the obstacles. 
This change in~$\theta$ is remarkable since the omnidirectional robot's rotational and translational kinematics are fully decoupled and the initial robot orientation is equal to the desired one. 
The middle plot of Fig.~\ref{fig:simulation} (top) shows the crucial overlap metric~$\kappa_i^{\star}$ for all robot-obstacle pairings $i\in\{1,2,3\}$, where $\kappa_i^\star$ is the expressions furnished for the respective obstacle as discussed in Sec.~\ref{sec:ellipsoid}. 
Additionally, the right plot of Fig.~\ref{fig:simulation} (top) displays the computation time needed to solve the underlying OCP~\eqref{eq:OCP} of the collision-avoidant MPC at each time instant.

Analogously, Fig.~\ref{fig:simulation} shows a simulation scenario for the differential-drive robot~\eqref{eq:kinematics_dd}. The robot's planar shape shall be an ellipsoid with semi-axis lengths of $\unit[0.2]{m}$ and $\unit[0.1]{m}$ yielding $\mathcal{E}_\textnormal{r}$ as previously.
Crucially, to pass the present obstacles~$\mathcal{E}_{\textnormal{c},i}$, $i\in\{1,2\}$, the robot needs to adapt its orientation not (only) for optimality reasons but also due to its nonholonomic kinematics.

\subsection{Hardware Experiments}
\begin{figure*}
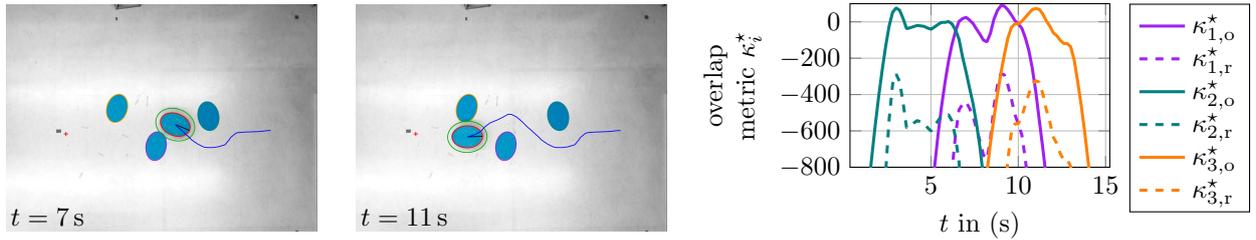

    \centering
    \include{figures/lab_png}
    \caption{Hardware results with still images for the omnidirectional mobile robot evading multiple obstacles. The crucial metric is shown on the right for each obstacle.}
    \label{fig:hardware}
\end{figure*}
Having shown the principal functionality of the proposed scheme using simulations, it remains to show that it is also applicable in practice using real-world hardware.
Figure~\ref{fig:hardware} exemplifies this where both the omnidirectional mobile robot and the obstacles are of some ellipsoidal shapes with the corresponding location vectors being tracked by a motion capture system.
For the robot, we use for collision avoidance an ellipse, depicted in green, that is slightly larger than the actual shape of the robot, which is marked by a red outline, to be able to deal with uncertainties. 
Starting at the initial pose $\bm{x}_0\approx\begin{bmatrix} \unit[2.87]{m} & \unit[0.01]{m} & \unit[0.02]{rad}\end{bmatrix}\tran$, the robot successfully evades all light-blue obstacles and again utilizes its rotational degree of freedom to squeeze through narrow passages in order to drive to the origin. 
The practical performance of the controller is additionally demonstrated in a video included as supplementary material of this paper.
The latter also comprises an additional hardware experiment and the simulative investigations presented earlier.
In the videos and in Fig.~\ref{fig:hardware}, the origin is indicated by a red cross, which is added to the image in post-processing for reference. 
Similarly, the obstacles' outlines have been marked with color during post-processing to be able to distinguish them, whereas the actual light-blue obstacles are physically present in the laboratory environment. 
The right-hand side plot of Fig.~\ref{fig:hardware} depicts the values of the collision-avoidance constraint functions $\kappa^\star_{i,j}$ over time, where the color as well as the first subscript $i\in\{1,2,3\}$ indicate the respective robot-obstacle-pairing and where the line-style as well as the second subscript $j$ indicate whether the metric is computed w.r.t.\ the green, outer ellipsoid (solid, $j=\textnormal{o}$) or w.r.t.\ the actual, red robot ellipsoid (dashed, $j=\textnormal{r}$).
Note that $\kappa^\star_{i,\textnormal{r}}$ is not included in the OCP but solely shown here to demonstrate that no collisions occur between the robot and the obstacles since $\kappa^\star_{i,\textnormal{r}}<0$ holds at all times for all robot-obstacle pairings~$i\in\lbrace 1,2,3\rbrace$.
However, for some time instances, the constraint~\eqref{eq:MPC_condition} is not strictly satisfied for the current measurement of the robot's pose, demonstrating one of the main difficulties when transitioning from simulation to hardware when using state constraints.
Plant-model mismatches, time-delays, and measurement inaccuracies may cause the ellipsoids to either indeed slightly overlap or to be misclassified as doing so. 
As done here, in practice, an actual collision can still be prevented by introducing additional safety margins by enlarging one or both of the involved ellipsoids as much as necessary for the accuracies obtainable with the hardware available. 
The needed enlargement is a function of the time delay, measurement inaccuracies, the sampling rate, the plant-model mismatch, and the maximum velocity of the robot. 
However, if a pose with a slight overlap is admitted while driving, the OCP becomes technically infeasible. 
In our case, it was sufficient to change the solver's acceptable constraint violation tolerance to a coarser value than the desired tolerance. 
However, if the system dynamics is incorporated into the problem via constraints, this can lead to violations of the dynamic equations if only one tolerance can be configured for all constraints. 
Thus, in contrast to our simulative investigations, in the hardware experiments, we used a condensed formulation of the OCP, i.e., the system dynamics is not represented as constraints but is inserted recursively into the optimization problem so that the inputs are the only free variables. 
However, as usual for any kind of state constraint, in more difficult and non-condensed cases, a more thorough treatment, in conjunction with the precise workings of the numerical solution algorithm, can be necessary. 

\section{Conclusions and Outlook}
\label{sec:conclusions}
This paper presents a collision-avoiding MPC approach for wheeled mobile robots, specifically dealing with arbitrary ellipsoidal shapes for both the robot and the obstacles.
Therefore, an overlap-like metric derived from an overlap test between two ellipsoids is novelly integrated into the underlying OCP. 
This yields a computationally efficient formulation of the typically complex task.
The effectiveness of the proposed control scheme has been demonstrated for two distinct kinematics in simulations. Hardware experiments have shown that collision avoidance is also achievable with plant-model mismatches and time delays as they realistically occur with real-world hardware.
Future work may focus on accelerating the OCP solution even more, which is readily achievable using recent advancements in numerical solvers~\citep{Vanroye23} and using fully compiled code. 
Moreover, one may combine the presented approach with machine-learned hotstarts of the optimization problem and with learned terminal costs to obtain long-term optimal trajectories even with short prediction horizons~\citep{Celestini24}.
Finally, one can seek to apply the concept to other types of robots, such as drones navigating around three-dimensional obstacles, as it is, although not yet tested experimentally, already covered by the approach.

\section{Acknowledgments}
The ITM acknowledges the support by the Deutsche Forschungsgemeinschaft (DFG, German Research Foundation) under Germany’s Excellence Strategy – EXC 2075 – 390740016, project PN4-4 “Learning from Data - Predictive Control in Adaptive Multi-Agent Scenarios” as well as project EB195/32-1, 433183605 “Research on Multibody Dynamics and Control for Collaborative Elastic Object Transportation by a Heterogeneous Swarm with Aerial and Land-Based Mobile Robots” and project EB195/40-1, 501890093 “Mehr Intelligenz wagen - Designassistenten in Mechanik und Dynamik (SPP 2353)”.\\
K.~Flaßkamp also acknowledges the support by the Deutsche Forschungsgemeinschaft (DFG, German Research Foundation) in project 501928699 “Intelligent design assistance for personalized medical surgery based on concentric tube continuum robots” within SPP 2353.

\bibliographystyle{elsarticle-num}
\bibliography{references}

\end{document}

%% file: figures/header_plots.tex
\definecolor{Kplot}{RGB}{255,165,0}
\definecolor{Kqplot}{RGB}{160,32,240}

\definecolor{obstacleSim}{RGB}{0,0,255}
\definecolor{robotSim}{RGB}{255,0,0}

\pgfplotsset{
    AxisEllipsoid/.style={
        width = 4cm,
        height = 4cm,
        ticks = none,
        axis equal,
    }
}
\def\xshiftAxisEll{.15cm}
\def\yshiftAxisEll{-.25cm}

\pgfplotsset{
    PlotEllipsoid/.style={
        mark = none,
        domain=0:360,
        samples=100,
        line width = .2mm,
        fill opacity = 0.5
    }
}

\definecolor{EllA}{RGB}{103,169,207}
\definecolor{EllB}{RGB}{33,102,172}
\definecolor{EllC}{RGB}{255,0,0}

\pgfplotsset{
    PlotEllipsoidA/.style={
        PlotEllipsoid,
        draw = EllA, 
        fill = EllA,
    }
}

\pgfplotsset{
    PlotEllipsoidB/.style={
        PlotEllipsoid,
        draw = EllB, 
        fill = EllB,
    }
}

\pgfplotsset{
    PlotEllipsoidC/.style={
        PlotEllipsoid,
        draw = EllC, 
        fill = EllC,
    }
}

\pgfplotsset{
    PlotEllipsoidSim/.style={
        PlotEllipsoid,
        fill opacity = 0.2
    }
}

\pgfplotsset{
    PlotMcenter/.style={
        color = black, 
        line width = 0.2mm
    }
}

\pgfplotsset{
    AxisEllipsoidKq/.style={
        width = 5cm,
        height = 4cm,
        ticks = none,
        axis equal,
    }
}
\pgfplotsset{
    AxisKq/.style={
        width = 5cm,
        height = 5cm,
        grid = major,
        xlabel={$\lambda$},
        xmin=0,
        xmax=1,
        ymin=-2.5,
        xtick distance=0.5,
    }
}
\def\xshiftAxisEllKq{.75cm}
\def\yshiftAxisEllKq{-.1cm}

\pgfplotsset{
    PlotEllipsoidKq/.style={
        mark = none,
        domain=0:360,
        samples=100,
        line width = .2mm,
        fill opacity = 0.2
    }
}
\def\linewidthKqplot{.4mm}
\pgfplotsset{
    PlotK/.style={
        mark = none,
        domain=0:1,
        samples=100,
        line width = \linewidthKqplot,
        draw = Kplot
    }
}
\pgfplotsset{
    PlotKq/.style={
        mark = none,
        domain=0:1,
        samples=100,
        line width = \linewidthKqplot,
        draw = Kqplot,
        dashed
    }
}
\pgfplotsset{
    PlotKqZero/.style={
        gray, 
        line width = 0.3mm,
    }
}

\pgfplotsset{
    PlotRobot/.style={
        mark = none,
        domain=0:360,
        samples=100,
        line width = .2mm,
    }
}

\tikzset{
    orientationSim/.style={
        line width = 0.2mm,
        draw = black,
        draw opacity = 0.8,
    }
}

%% file: figures/figure_ellipsoids.tex
\begin{tikzpicture}[every node/.style={inner sep=0}]

    \begin{axis}[
        AxisEllipsoid,
        name=lambdaA   
        ]
        \readlist\pEllA{1.5,.8,-0.8,0,45} 
        \readlist\pEllB{1,0.4, 0.4, -0.6, 0} 
        \readlist\pEllC{0.8748, 0.3655, 0.2473, -0.5793, -179.3222} 
        \addplot+[PlotEllipsoidA] 
            ({(\pEllA[1]* cos(x))*cos(\pEllA[5]) - (\pEllA[2] * sin(x))*sin(\pEllA[5]) + \pEllA[3]}, {(\pEllA[1] * cos(x))*sin(\pEllA[5]) + (\pEllA[2] * sin(x))*cos(\pEllA[5])} +\pEllA[4]); 
        \addplot+[PlotEllipsoidB] 
            ({(\pEllB[1]* cos(x))*cos(\pEllB[5]) - (\pEllB[2] * sin(x))*sin(\pEllB[5]) + \pEllB[3]}, {(\pEllB[1] * cos(x))*sin(\pEllB[5]) + (\pEllB[2] * sin(x))*cos(\pEllB[5])} +\pEllB[4]); 
        \addplot[PlotMcenter] table[x index=3, y index=4]{src/data/data_ell_20241114_152154.txt};   
        \addplot+[PlotEllipsoidC] 
            ({(\pEllC[1]* cos(x))*cos(\pEllC[5]) - (\pEllC[2] * sin(x))*sin(\pEllC[5]) + \pEllC[3]}, {(\pEllC[1] * cos(x))*sin(\pEllC[5]) + (\pEllC[2] * sin(x))*cos(\pEllC[5])} +\pEllC[4]); 
        \node[EllC] at (\pEllC[3], \pEllC[4]) {\scriptsize{\textbullet}};
        \node[EllA] at (\pEllA[3], \pEllA[4]) {\scriptsize{\textbullet}};
        \node[EllB] at (\pEllB[3], \pEllB[4]) {\scriptsize{\textbullet}};
        \node[anchor = north west] at (rel axis cs: 0.01, 0.99) {{$\lambda=0.1$}}; 
        \node[anchor = south east, EllC] at (rel axis cs: 0.99, 0.01) {\scriptsize{$K(\lambda)=0.77$}}; 
    \end{axis}

    \begin{axis}[
        AxisEllipsoid,
        name=lambdaB,
        at={(lambdaA.east)},
        anchor=west,
        xshift = \xshiftAxisEll,
        ]
        \readlist\pEllA{1.5,.8,-0.8,0,45} 
        \readlist\pEllB{1,0.4, 0.4, -0.6, 0} 
        \readlist\pEllC{0.6728, 0.3099, -0.0504, -0.5345, -177.3925} 
        \addplot+[PlotEllipsoidA] 
            ({(\pEllA[1]* cos(x))*cos(\pEllA[5]) - (\pEllA[2] * sin(x))*sin(\pEllA[5]) + \pEllA[3]}, {(\pEllA[1] * cos(x))*sin(\pEllA[5]) + (\pEllA[2] * sin(x))*cos(\pEllA[5])} +\pEllA[4]); 
        \addplot+[PlotEllipsoidB] 
            ({(\pEllB[1]* cos(x))*cos(\pEllB[5]) - (\pEllB[2] * sin(x))*sin(\pEllB[5]) + \pEllB[3]}, {(\pEllB[1] * cos(x))*sin(\pEllB[5]) + (\pEllB[2] * sin(x))*cos(\pEllB[5])} +\pEllB[4]); 
        \addplot[PlotMcenter] table[x index=3, y index=4]{src/data/data_ell_20241114_152154.txt};  
        \addplot+[PlotEllipsoidC] 
            ({(\pEllC[1]* cos(x))*cos(\pEllC[5]) - (\pEllC[2] * sin(x))*sin(\pEllC[5]) + \pEllC[3]}, {(\pEllC[1] * cos(x))*sin(\pEllC[5]) + (\pEllC[2] * sin(x))*cos(\pEllC[5])} +\pEllC[4]); 
        \node[EllC] at (\pEllC[3], \pEllC[4]) {\scriptsize{\textbullet}};
        \node[EllA] at (\pEllA[3], \pEllA[4]) {\scriptsize{\textbullet}};
        \node[EllB] at (\pEllB[3], \pEllB[4]) {\scriptsize{\textbullet}};
        \node[anchor = north west] at (rel axis cs: 0.01, 0.99) {{$\lambda=0.3$}}; 
        \node[anchor = south east, EllC] at (rel axis cs: 0.99, 0.01) {\scriptsize{$K(\lambda)=0.45$}}; 
    \end{axis}

    \begin{axis}[
        AxisEllipsoid,
        name=lambdaG,
        at={(lambdaB.east)},
        anchor=west,
        xshift = \xshiftAxisEll,
        ]
        \readlist\pEllA{1.5,.8,-0.8,0,45} 
        \readlist\pEllB{1,0.4, 0.4, -0.6, 0} 
        \readlist\pEllC{0.587878940049565, 0.303162579259786, -0.335303095049142, -0.481179415176119, -173.988861656908} 
        \addplot+[PlotEllipsoidA] 
            ({(\pEllA[1]* cos(x))*cos(\pEllA[5]) - (\pEllA[2] * sin(x))*sin(\pEllA[5]) + \pEllA[3]}, {(\pEllA[1] * cos(x))*sin(\pEllA[5]) + (\pEllA[2] * sin(x))*cos(\pEllA[5])} +\pEllA[4]); 
        \addplot+[PlotEllipsoidB] 
            ({(\pEllB[1]* cos(x))*cos(\pEllB[5]) - (\pEllB[2] * sin(x))*sin(\pEllB[5]) + \pEllB[3]}, {(\pEllB[1] * cos(x))*sin(\pEllB[5]) + (\pEllB[2] * sin(x))*cos(\pEllB[5])} +\pEllB[4]); 
        \addplot[PlotMcenter] table[x index=3, y index=4]{src/data/data_ell_20241114_152154.txt};  
        \addplot+[PlotEllipsoidC] 
            ({(\pEllC[1]* cos(x))*cos(\pEllC[5]) - (\pEllC[2] * sin(x))*sin(\pEllC[5]) + \pEllC[3]}, {(\pEllC[1] * cos(x))*sin(\pEllC[5]) + (\pEllC[2] * sin(x))*cos(\pEllC[5])} +\pEllC[4]); 
        \node[EllC] at (\pEllC[3], \pEllC[4]) {\scriptsize{\textbullet}};
        \node[EllA] at (\pEllA[3], \pEllA[4]) {\scriptsize{\textbullet}};
        \node[EllB] at (\pEllB[3], \pEllB[4]) {\scriptsize{\textbullet}};
        \node[anchor = north west] at (rel axis cs: 0.01, 0.99) {{$\lambda=0.5$}}; 
        \node[anchor = south east, EllC] at (rel axis cs: 0.99, 0.01) {\scriptsize{$K(\lambda)=0.34$}}; 
    \end{axis}

    \begin{axis}[
        AxisEllipsoid,
        name=lambdaH,
        at={(lambdaG.east)},
        anchor=west,
        xshift = \xshiftAxisEll,
        ]
        \readlist\pEllA{1.5,.8,-0.8,0,45} 
        \readlist\pEllB{1,0.4, 0.4, -0.6, 0} 
        \readlist\pEllC{0.680770899124136, 0.397610155369454, -0.599371430816318, -0.40602069278324, -166.788297233202} 
        \addplot+[PlotEllipsoidA] 
            ({(\pEllA[1]* cos(x))*cos(\pEllA[5]) - (\pEllA[2] * sin(x))*sin(\pEllA[5]) + \pEllA[3]}, {(\pEllA[1] * cos(x))*sin(\pEllA[5]) + (\pEllA[2] * sin(x))*cos(\pEllA[5])} +\pEllA[4]); 
        \addplot+[PlotEllipsoidB] 
            ({(\pEllB[1]* cos(x))*cos(\pEllB[5]) - (\pEllB[2] * sin(x))*sin(\pEllB[5]) + \pEllB[3]}, {(\pEllB[1] * cos(x))*sin(\pEllB[5]) + (\pEllB[2] * sin(x))*cos(\pEllB[5])} +\pEllB[4]); 
        \addplot[PlotMcenter] table[x index=3, y index=4]{src/data/data_ell_20241114_152154.txt};   
        \addplot+[PlotEllipsoidC] 
            ({(\pEllC[1]* cos(x))*cos(\pEllC[5]) - (\pEllC[2] * sin(x))*sin(\pEllC[5]) + \pEllC[3]}, {(\pEllC[1] * cos(x))*sin(\pEllC[5]) + (\pEllC[2] * sin(x))*cos(\pEllC[5])} +\pEllC[4]); 
        \node[EllC] at (\pEllC[3], \pEllC[4]) {\scriptsize{\textbullet}};
        \node[EllA] at (\pEllA[3], \pEllA[4]) {\scriptsize{\textbullet}};
        \node[EllB] at (\pEllB[3], \pEllB[4]) {\scriptsize{\textbullet}};
        \node[anchor = north west] at (rel axis cs: 0.01, 0.99) {{$\lambda=0.6$}}; 
        \node[anchor = south east, EllC] at (rel axis cs: 0.99, 0.01) {\scriptsize{$K(\lambda)=0.42$}}; 
    \end{axis}

    \begin{axis}[
        AxisEllipsoid,
        name=lambdaI,
        at={(lambdaH.east)},
        anchor=west,
        xshift = \xshiftAxisEll,
        ]
        \readlist\pEllA{1.5,.8,-0.8,0,45} 
        \readlist\pEllB{1,0.4, 0.4, -0.6, 0} 
        \readlist\pEllC{1.0177, 0.6280, -0.8041, -0.2468, -148.7764} 
        \addplot+[PlotEllipsoidA] 
            ({(\pEllA[1]* cos(x))*cos(\pEllA[5]) - (\pEllA[2] * sin(x))*sin(\pEllA[5]) + \pEllA[3]}, {(\pEllA[1] * cos(x))*sin(\pEllA[5]) + (\pEllA[2] * sin(x))*cos(\pEllA[5])} +\pEllA[4]); 
        \addplot+[PlotEllipsoidB] 
            ({(\pEllB[1]* cos(x))*cos(\pEllB[5]) - (\pEllB[2] * sin(x))*sin(\pEllB[5]) + \pEllB[3]}, {(\pEllB[1] * cos(x))*sin(\pEllB[5]) + (\pEllB[2] * sin(x))*cos(\pEllB[5])} +\pEllB[4]); 
        \addplot[PlotMcenter] table[x index=3, y index=4]{src/data/data_ell_20241114_152154.txt};   
        \addplot[PlotEllipsoidC] 
            ({(\pEllC[1]* cos(x))*cos(\pEllC[5]) - (\pEllC[2] * sin(x))*sin(\pEllC[5]) + \pEllC[3]}, {(\pEllC[1] * cos(x))*sin(\pEllC[5]) + (\pEllC[2] * sin(x))*cos(\pEllC[5])} +\pEllC[4]); 
        \node[EllC] at (\pEllC[3], \pEllC[4]) {\scriptsize{\textbullet}};
        \node[EllA] at (\pEllA[3], \pEllA[4]) {\scriptsize{\textbullet}};
        \node[EllB] at (\pEllB[3], \pEllB[4]) {\scriptsize{\textbullet}};
        \node[anchor = north west] at (rel axis cs: 0.01, 0.99) {{$\lambda=0.9$}}; 
        \node[anchor = south east, EllC] at (rel axis cs: 0.99, 0.01) {\scriptsize{$K(\lambda)=0.72$}}; 
    \end{axis}

    \begin{axis}[
        AxisEllipsoid,
        name=lambdaAb,  
        at={(lambdaA.south west)}, 
        anchor = north west,
        yshift= \yshiftAxisEll,
        ]
        \readlist\pEllA{1.5,.8,-0.8,0,45} 
        \readlist\pEllB{1,0.4, 0.4, -1.4, 0} 
        \readlist\pEllC{0.7193, 0.3005, 0.2033, -1.3657, -179.3222} 
        \addplot+[PlotEllipsoidA] 
            ({(\pEllA[1]* cos(x))*cos(\pEllA[5]) - (\pEllA[2] * sin(x))*sin(\pEllA[5]) + \pEllA[3]}, {(\pEllA[1] * cos(x))*sin(\pEllA[5]) + (\pEllA[2] * sin(x))*cos(\pEllA[5])} +\pEllA[4]); 
        \addplot+[PlotEllipsoidB] 
            ({(\pEllB[1]* cos(x))*cos(\pEllB[5]) - (\pEllB[2] * sin(x))*sin(\pEllB[5]) + \pEllB[3]}, {(\pEllB[1] * cos(x))*sin(\pEllB[5]) + (\pEllB[2] * sin(x))*cos(\pEllB[5])} +\pEllB[4]); 
        \addplot[PlotMcenter] table[x index=3, y index=4]{src/data/data_ell_20241114_170858.txt};   
        \addplot[PlotEllipsoidC] 
            ({(\pEllC[1]* cos(x))*cos(\pEllC[5]) - (\pEllC[2] * sin(x))*sin(\pEllC[5]) + \pEllC[3]}, {(\pEllC[1] * cos(x))*sin(\pEllC[5]) + (\pEllC[2] * sin(x))*cos(\pEllC[5])} +\pEllC[4]); 
        \node[EllC] at (\pEllC[3], \pEllC[4]) {\scriptsize{\textbullet}};
        \node[EllA] at (\pEllA[3], \pEllA[4]) {\scriptsize{\textbullet}};
        \node[EllB] at (\pEllB[3], \pEllB[4]) {\scriptsize{\textbullet}};
        \node[anchor = north west] at (rel axis cs: 0.01, 0.99) {{$\lambda=0.1$}}; 
        \node[anchor = south east, EllC] at (rel axis cs: 0.99, 0.01) {\scriptsize{$K(\lambda)=0.52$}}; 
    \end{axis}

    \begin{axis}[
        AxisEllipsoid,
        name=lambdaBb,
        at={(lambdaAb.east)},
        anchor=west,
        xshift = \xshiftAxisEll,
        ]
        \readlist\pEllA{1.5,.8,-0.8,0,45} 
        \readlist\pEllB{1,0.4, 0.4, -1.4, 0} 
        \readlist\pEllC{0.3557, 0.1558, 0.0109, -1.3285, -178.4762} 
        \addplot+[PlotEllipsoidA] 
            ({(\pEllA[1]* cos(x))*cos(\pEllA[5]) - (\pEllA[2] * sin(x))*sin(\pEllA[5]) + \pEllA[3]}, {(\pEllA[1] * cos(x))*sin(\pEllA[5]) + (\pEllA[2] * sin(x))*cos(\pEllA[5])} +\pEllA[4]); 
        \addplot+[PlotEllipsoidB] 
            ({(\pEllB[1]* cos(x))*cos(\pEllB[5]) - (\pEllB[2] * sin(x))*sin(\pEllB[5]) + \pEllB[3]}, {(\pEllB[1] * cos(x))*sin(\pEllB[5]) + (\pEllB[2] * sin(x))*cos(\pEllB[5])} +\pEllB[4]); 
        \addplot[PlotMcenter] table[x index=3, y index=4]{src/data/data_ell_20241114_170858.txt};     
        \addplot[PlotEllipsoidC] 
            ({(\pEllC[1]* cos(x))*cos(\pEllC[5]) - (\pEllC[2] * sin(x))*sin(\pEllC[5]) + \pEllC[3]}, {(\pEllC[1] * cos(x))*sin(\pEllC[5]) + (\pEllC[2] * sin(x))*cos(\pEllC[5])} +\pEllC[4]); 
        \node[EllC] at (\pEllC[3], \pEllC[4]) {\scriptsize{\textbullet}};
        \node[EllA] at (\pEllA[3], \pEllA[4]) {\scriptsize{\textbullet}};
        \node[EllB] at (\pEllB[3], \pEllB[4]) {\scriptsize{\textbullet}};
        \node[anchor = north west] at (rel axis cs: 0.01, 0.99) {{$\lambda=0.2$}}; 
        \node[anchor = south east, EllC] at (rel axis cs: 0.99, 0.01) {\scriptsize{$K(\lambda)=0.13$}}; 
    \end{axis}

    \begin{axis}[
        AxisEllipsoid,
        name=lambdaBbb,
        at={(lambdaBb.east)},
        anchor=west,
        xshift = \xshiftAxisEll,
        ]
        \readlist\pEllA{1.5,.8,-0.8,0,45} 
        \readlist\pEllB{1,0.4, 0.4, -1.4, 0} 
        \readlist\pEllC{0,0,-0.531872291068921,-1.18565411147061,6.01113834309176} 
        \addplot+[PlotEllipsoidA] 
            ({(\pEllA[1]* cos(x))*cos(\pEllA[5]) - (\pEllA[2] * sin(x))*sin(\pEllA[5]) + \pEllA[3]}, {(\pEllA[1] * cos(x))*sin(\pEllA[5]) + (\pEllA[2] * sin(x))*cos(\pEllA[5])} +\pEllA[4]); 
        \addplot+[PlotEllipsoidB] 
            ({(\pEllB[1]* cos(x))*cos(\pEllB[5]) - (\pEllB[2] * sin(x))*sin(\pEllB[5]) + \pEllB[3]}, {(\pEllB[1] * cos(x))*sin(\pEllB[5]) + (\pEllB[2] * sin(x))*cos(\pEllB[5])} +\pEllB[4]); 
        \addplot[PlotMcenter] table[x index=3, y index=4]{src/data/data_ell_20241114_170858.txt};     
        \addplot[PlotEllipsoidC] 
            ({(\pEllC[1]* cos(x))*cos(\pEllC[5]) - (\pEllC[2] * sin(x))*sin(\pEllC[5]) + \pEllC[3]}, {(\pEllC[1] * cos(x))*sin(\pEllC[5]) + (\pEllC[2] * sin(x))*cos(\pEllC[5])} +\pEllC[4]); 
        \node[EllC] at (\pEllC[3], \pEllC[4]) {\scriptsize{\textbullet}};
        \node[EllA] at (\pEllA[3], \pEllA[4]) {\scriptsize{\textbullet}};
        \node[EllB] at (\pEllB[3], \pEllB[4]) {\scriptsize{\textbullet}};
        \node[anchor = north west] at (rel axis cs: 0.01, 0.99) {{$\lambda=0.5$}}; 
        \node[anchor = south east, EllC] at (rel axis cs: 0.99, 0.01) {\scriptsize{$K(\lambda)=-0.50$}}; 
    \end{axis}

    \begin{axis}[
        AxisEllipsoid,
        name=lambdaCb,
        at={(lambdaBbb.east)},
        anchor=west,
        xshift = \xshiftAxisEll,
        ]
        \readlist\pEllA{1.5,.8,-0.8,0,45} 
        \readlist\pEllB{1,0.4, 0.4, -1.4, 0} 
        \readlist\pEllC{0,0,-0.956027202750168,-0.886495777595097	, 0.2130393700496} 
        \addplot+[PlotEllipsoidA] 
            ({(\pEllA[1]* cos(x))*cos(\pEllA[5]) - (\pEllA[2] * sin(x))*sin(\pEllA[5]) + \pEllA[3]}, {(\pEllA[1] * cos(x))*sin(\pEllA[5]) + (\pEllA[2] * sin(x))*cos(\pEllA[5])} +\pEllA[4]); 
        \addplot+[PlotEllipsoidB] 
            ({(\pEllB[1]* cos(x))*cos(\pEllB[5]) - (\pEllB[2] * sin(x))*sin(\pEllB[5]) + \pEllB[3]}, {(\pEllB[1] * cos(x))*sin(\pEllB[5]) + (\pEllB[2] * sin(x))*cos(\pEllB[5])} +\pEllB[4]); 
        \addplot[PlotMcenter] table[x index=3, y index=4]{src/data/data_ell_20241114_170858.txt};      
        \node[EllC] at (\pEllC[3], \pEllC[4]) {\scriptsize{\textbullet}};  
        \node[EllA] at (\pEllA[3], \pEllA[4]) {\scriptsize{\textbullet}};
        \node[EllB] at (\pEllB[3], \pEllB[4]) {\scriptsize{\textbullet}};
        \node[anchor = north west] at (rel axis cs: 0.01, 0.99) {{$\lambda=0.8$}}; 
        \node[anchor = south east, EllC] at (rel axis cs: 0.99, 0.01) {\scriptsize{$K(\lambda)=-0.22$}}; 
    \end{axis}

    \begin{axis}[
        AxisEllipsoid,
        name=lambdaDb,
        at={(lambdaCb.east)},
        anchor=west,
        xshift = \xshiftAxisEll,
        ]
        \readlist\pEllA{1.5,.8,-0.8,0,45} 
        \readlist\pEllB{1,0.4, 0.4, -1.4, 0} 
        \readlist\pEllC{0.4873, 0.3007, -1.0007, -0.6387, -148.7764} 
        \addplot+[PlotEllipsoidA] 
            ({(\pEllA[1]* cos(x))*cos(\pEllA[5]) - (\pEllA[2] * sin(x))*sin(\pEllA[5]) + \pEllA[3]}, {(\pEllA[1] * cos(x))*sin(\pEllA[5]) + (\pEllA[2] * sin(x))*cos(\pEllA[5])} +\pEllA[4]); 
        \addplot+[PlotEllipsoidB] 
            ({(\pEllB[1]* cos(x))*cos(\pEllB[5]) - (\pEllB[2] * sin(x))*sin(\pEllB[5]) + \pEllB[3]}, {(\pEllB[1] * cos(x))*sin(\pEllB[5]) + (\pEllB[2] * sin(x))*cos(\pEllB[5])} +\pEllB[4]); 
        \addplot[PlotMcenter] table[x index=3, y index=4]{src/data/data_ell_20241114_170858.txt}; 
        \addplot[PlotEllipsoidC] 
            ({(\pEllC[1]* cos(x))*cos(\pEllC[5]) - (\pEllC[2] * sin(x))*sin(\pEllC[5]) + \pEllC[3]}, {(\pEllC[1] * cos(x))*sin(\pEllC[5]) + (\pEllC[2] * sin(x))*cos(\pEllC[5])} +\pEllC[4]); 
        \node[EllC] at (\pEllC[3], \pEllC[4]) {\scriptsize{\textbullet}};
        \node[EllA] at (\pEllA[3], \pEllA[4]) {\scriptsize{\textbullet}};
        \node[EllB] at (\pEllB[3], \pEllB[4]) {\scriptsize{\textbullet}};
        \node[anchor = north west] at (rel axis cs: 0.01, 0.99) {{$\lambda=0.9$}}; 
        \node[anchor = south east, EllC] at (rel axis cs: 0.99, 0.01) {\scriptsize{$K(\lambda)=0.17$}}; 
    \end{axis}

\end{tikzpicture}

%% file: figures/figure_K.tex
\begin{tikzpicture}%
    \begin{axis}[
        AxisKq,
        name=KCaseA,
        anchor = north west,
        ]
        \addplot[PlotK] table[x index=0, y index=1]{src/data/Kq_data_ell_20241115_084038.txt};  
        \addplot[PlotKq] table[x index=0, y index=2]{src/data/Kq_data_ell_20241115_084038.txt};  
        \addplot[PlotKqZero] {0};
        \node[anchor=north east, fill = white, inner sep=0pt, outer sep=0pt] at (rel axis cs: 1,1) {
            \includegraphics[width=2cm]{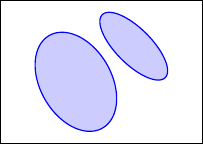}};
 \end{axis}%
    \begin{axis}[
        AxisKq,
        name=KCaseB,
        at={(KCaseA.east)}, 
        anchor = west,
        xshift = \xshiftAxisEllKq,
        legend cell align={left},
        legend columns = 1,
        legend style={
          fill=white,
          at={(1.0,1.0)},
          anchor=north east,
          column sep = 0.1cm,
        }
        ]
        \addplot[PlotK] table[x index=0, y index=1]{src/data/Kq_data_ell_20241115_083835.txt};
        \addplot[PlotKq] table[x index=0, y index=2]{src/data/Kq_data_ell_20241115_083835.txt};
        \addplot[PlotKqZero] {0};
        \node[anchor=north east, fill = white, inner sep=0pt, outer sep=0pt] at (rel axis cs: 1,1) {
            \includegraphics[width=2cm]{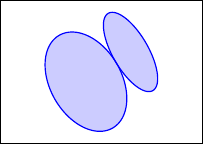}};
\end{axis}%
    \begin{axis}[
        AxisKq,
        name=KCaseC,
        at={(KCaseB.east)}, 
        anchor = west,
        xshift = \xshiftAxisEllKq,
        legend cell align={left},
        legend columns = 1,
        legend style={
          fill=white,
          at={(1.05,0.5)},
          anchor= west,
          column sep = 0.1cm,
        }
        ]
        \addplot[PlotK] table[x index=0, y index=1]{src/data/Kq_data_ell_20241115_084001.txt};  \addlegendentry{$K(\lambda)$} 
        \addplot[PlotKq] table[x index=0, y index=2]{src/data/Kq_data_ell_20241115_084001.txt};  \addlegendentry{$K(\lambda) q(\lambda)$} 
        \addplot[PlotKqZero] {0};
        \node[anchor=north east, fill = white, inner sep=0pt, outer sep=0pt] at (rel axis cs: 1,1) {
            \includegraphics[width=2cm]{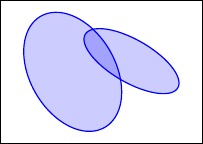}};
\end{axis}%
\end{tikzpicture}%

%% file: figures/simulation_fig.tex
\def\heightAll{4.5cm}
\def\widthKq{6cm}
\def\widthT{4.25cm}

\begin{tikzpicture}

    \begin{axis}[
        name = simulationPlane,
        height=\heightAll,
        xmajorgrids, 
        ymajorgrids,
        axis equal,
        ylabel = {$x_2$ $(\unit{m})$}
        ]	

        \readlist\pEllA{1.0,0.1,0.0,-0.3,0} 
        \readlist\pEllB{0.075, 0.3,-0.45,-0.15,0} 
        \readlist\pEllC{0.075, 0.3,0.45,-0.15,0} 
        
        \addplot[PlotEllipsoidSim, draw = Kqplot, fill = obstacleSim] 
            ({(\pEllA[1]* cos(x))*cos(\pEllA[5]) - (\pEllA[2] * sin(x))*sin(\pEllA[5]) + \pEllA[3]}, {(\pEllA[1] * cos(x))*sin(\pEllA[5]) + (\pEllA[2] * sin(x))*cos(\pEllA[5])} +\pEllA[4]); 
        \addplot[PlotEllipsoidSim, draw = teal, fill = obstacleSim] 
            ({(\pEllB[1]* cos(x))*cos(\pEllB[5]) - (\pEllB[2] * sin(x))*sin(\pEllB[5]) + \pEllB[3]}, {(\pEllB[1] * cos(x))*sin(\pEllB[5]) + (\pEllB[2] * sin(x))*cos(\pEllB[5])} +\pEllB[4]); 
        \addplot[PlotEllipsoidSim, draw = orange, fill = obstacleSim] 
            ({(\pEllC[1]* cos(x))*cos(\pEllC[5]) - (\pEllC[2] * sin(x))*sin(\pEllC[5]) + \pEllC[3]}, {(\pEllC[1] * cos(x))*sin(\pEllC[5]) + (\pEllC[2] * sin(x))*cos(\pEllC[5])} +\pEllC[4]); 

        \addplot[color = blue, line width = 0.4mm, each nth point=1] table[x index=1,y index=2]{src/data/SimExample/x_CL_20241120_173047.txt}; 

            \readlist\robPa{0.35,0.2,-1, 0.4,0} 
            \addplot[PlotRobot, draw = robotSim, fill = robotSim, fill opacity = 0.2, draw opacity = 0.7] 
                ({(\robPa[1]* cos(x))*cos(\robPa[5]) - (\robPa[2] * sin(x))*sin(\robPa[5]) +\robPa[3]}, {(\robPa[1] * cos(x))*sin(\robPa[5]) + (\robPa[2] * sin(x))*cos(\robPa[5]) +\robPa[4]});
            \pgfmathsetmacro{\OrientXa}{\robPa[3] + \robPa[1]*cos(\robPa[5])}
            \pgfmathsetmacro{\OrientYa}{\robPa[4] + \robPa[1]*sin(\robPa[5])}
            \draw[orientationSim, draw opacity = 0.7] (\robPa[3],\robPa[4]]) -- (\OrientXa,\OrientYa) node[robotSim, xshift=-4mm, above, yshift = 3mm] {\footnotesize{$t=\unit[0]{s}$}};  
            \readlist\robPb{0.35, 0.2,-0.201, 0.23974, -19.664} 
            \addplot[PlotRobot, draw = robotSim, fill = robotSim, fill opacity = 0.4, draw opacity = 0.85] 
                ({(\robPb[1]* cos(x))*cos(\robPb[5]) - (\robPb[2] * sin(x))*sin(\robPb[5]) +\robPb[3]}, {(\robPb[1] * cos(x))*sin(\robPb[5]) + (\robPb[2] * sin(x))*cos(\robPb[5]) +\robPb[4]});
            \pgfmathsetmacro{\OrientXb}{\robPb[3] + \robPb[1]*cos(\robPb[5])}
            \pgfmathsetmacro{\OrientYb}{\robPb[4] + \robPb[1]*sin(\robPb[5])}
            \draw[orientationSim, draw opacity = 0.85] (\robPb[3],\robPb[4]]) -- (\OrientXb,\OrientYb) node[robotSim, above, yshift = 4mm] {\footnotesize{$t=\unit[4]{s}$}};  
            \readlist\robPc{0.35, 0.2, 0,0,0} 
            \addplot[PlotRobot, draw = robotSim, fill = robotSim, fill opacity = 0.6, draw opacity = 1.0] 
                ({(\robPc[1]* cos(x))*cos(\robPc[5]) - (\robPc[2] * sin(x))*sin(\robPc[5]) +\robPc[3]}, {(\robPc[1] * cos(x))*sin(\robPc[5]) + (\robPc[2] * sin(x))*cos(\robPc[5]) +\robPc[4]});
            \pgfmathsetmacro{\OrientXc}{\robPc[3] + \robPc[1]*cos(\robPc[5])}
            \pgfmathsetmacro{\OrientYc}{\robPc[4] + \robPc[1]*sin(\robPc[5])}
            \draw[orientationSim, draw opacity = 1.0] (\robPc[3],\robPc[4]]) -- (\OrientXc,\OrientYc); 
    \end{axis}

    \begin{axis}[
        name=simulationKlambda,
        at={(simulationPlane.south east)},
        anchor=south west,
        xshift = 2cm,
        height=\heightAll,
        width = \widthKq,
        xmajorgrids, 
        ymajorgrids,
        restrict y to domain=-900:20,
        xmin=0,
        xmax=10,
        ymin=-400,
        ylabel={overlap metric $\kappa^\star_i$}, 
        legend cell align={left},
        legend columns = 1,
        legend style={
            at={(1.0,0.0)},
            anchor=south east,
            }
        ]
 
        \addplot[color = Kqplot, line width = 0.4mm, each nth point=1, line join=bevel] table[x index=0,y index=1]{src/data/SimExample/Kq_20241120_173047.txt}; \addlegendentry{$\kappa^{\star}_{1}$}
        \addplot[color = teal, line width = 0.4mm, each nth point=1, line join=bevel] table[x index=0,y index=2]{src/data/SimExample/Kq_20241120_173047.txt}; \addlegendentry{$\kappa^{\star}_{2}$}
        \addplot[color = orange, line width = 0.4mm, each nth point=1, line join=bevel] table[x index=0,y index=3]{src/data/SimExample/Kq_20241120_173047.txt}; \addlegendentry{$\kappa^{\star}_{2}$}
        
    \end{axis}

    \begin{axis}[
        name=simulationComptime,
        at={(simulationKlambda.south east)},
        anchor=south west,
        xshift = 1.75cm,
        height=\heightAll,
        width = \widthT,
        xmajorgrids, 
        ymajorgrids,
        ylabel={solution time (ms)},
        ymode=log,
        log ticks with fixed point, 
        ymin = 1,
        ymax = 500,
        xmin=0,
        xmax=10,
        ]
 
        \addplot[color = blue, only marks, mark = x, line width = 0.2mm, each nth point=1] table[x index=0, y expr=1e3*\thisrowno{1}]{src/data/SimExample/CompTime_20241120_173047.txt};
        \addplot[domain=0:20, samples=10, color=gray, thick] {200}; 
        \node[gray] at (7.25,290) {$\delta_t$};

    \end{axis}

    \begin{axis}[
        name = simulationPlaneDIANA,
        at={(simulationPlane.south west)},
        anchor=north west,
        yshift = -.75cm,
        height=\heightAll,
        xmajorgrids, 
        ymajorgrids,
        axis equal,
        xlabel = {$x_1$ $(\unit{m})$},
        ylabel = {$x_2$ $(\unit{m})$}
        ]	

        \readlist\pEllA{0.1000, 0.0500, 0.6000, 0.0500, 0} 
        \readlist\pEllB{0.1000, 0.0500, 0.3000, -0.2000, 90} 

        \addplot[PlotEllipsoidSim, draw = Kqplot, fill = obstacleSim] 
            ({(\pEllA[1]* cos(x))*cos(\pEllA[5]) - (\pEllA[2] * sin(x))*sin(\pEllA[5]) + \pEllA[3]}, {(\pEllA[1] * cos(x))*sin(\pEllA[5]) + (\pEllA[2] * sin(x))*cos(\pEllA[5])} +\pEllA[4]); 
        \addplot[PlotEllipsoidSim, draw = teal, fill = obstacleSim] 
            ({(\pEllB[1]* cos(x))*cos(\pEllB[5]) - (\pEllB[2] * sin(x))*sin(\pEllB[5]) + \pEllB[3]}, {(\pEllB[1] * cos(x))*sin(\pEllB[5]) + (\pEllB[2] * sin(x))*cos(\pEllB[5])} +\pEllB[4]); 

        \addplot[color = blue, line width = 0.4mm, each nth point=1] table[x index=1,y index=2]{src/data/SimExampleDIANA/x_CL_20241120_173405.txt}; 

            \readlist\robPa{0.2,0.1,1,0,0} 
            \addplot[PlotRobot, draw = robotSim, fill = robotSim, fill opacity = 0.2, draw opacity = 0.7] 
                ({(\robPa[1]* cos(x))*cos(\robPa[5]) - (\robPa[2] * sin(x))*sin(\robPa[5]) +\robPa[3]}, {(\robPa[1] * cos(x))*sin(\robPa[5]) + (\robPa[2] * sin(x))*cos(\robPa[5]) +\robPa[4]});
            \pgfmathsetmacro{\OrientXa}{\robPa[3] + \robPa[1]*cos(\robPa[5])}
            \pgfmathsetmacro{\OrientYa}{\robPa[4] + \robPa[1]*sin(\robPa[5])}
            \draw[orientationSim, draw opacity = 0.7] (\robPa[3],\robPa[4]]) -- (\OrientXa,\OrientYa) node[robotSim, midway, above, yshift = 2mm] {\footnotesize{$t=\unit[0]{s}$}};  
            \readlist\robPb{0.2,0.1,0.432114298841239, -0.0594563226915162, -27.7462832257} 
            \addplot[PlotRobot, draw = robotSim, fill = robotSim, fill opacity = 0.4, draw opacity = 0.85] 
                ({(\robPb[1]* cos(x))*cos(\robPb[5]) - (\robPb[2] * sin(x))*sin(\robPb[5]) +\robPb[3]}, {(\robPb[1] * cos(x))*sin(\robPb[5]) + (\robPb[2] * sin(x))*cos(\robPb[5]) +\robPb[4]});
            \pgfmathsetmacro{\OrientXb}{\robPb[3] + \robPb[1]*cos(\robPb[5])}
            \pgfmathsetmacro{\OrientYb}{\robPb[4] + \robPb[1]*sin(\robPb[5])}
            \draw[orientationSim, draw opacity = 0.85] (\robPb[3],\robPb[4]]) -- (\OrientXb,\OrientYb) node[robotSim, midway, below, yshift = -2mm, xshift=2mm] {\footnotesize{$t=\unit[2]{s}$}};  
            \readlist\robPc{0.2,0.1,0,0,0} 
            \addplot[PlotRobot, draw = robotSim, fill = robotSim, fill opacity = 0.6, draw opacity = 1.0] 
                ({(\robPc[1]* cos(x))*cos(\robPc[5]) - (\robPc[2] * sin(x))*sin(\robPc[5]) +\robPc[3]}, {(\robPc[1] * cos(x))*sin(\robPc[5]) + (\robPc[2] * sin(x))*cos(\robPc[5]) +\robPc[4]});
            \pgfmathsetmacro{\OrientXc}{\robPc[3] + \robPc[1]*cos(\robPc[5])}
            \pgfmathsetmacro{\OrientYc}{\robPc[4] + \robPc[1]*sin(\robPc[5])}
            \draw[orientationSim, draw opacity = 1.0] (\robPc[3],\robPc[4]]) -- (\OrientXc,\OrientYc); 
    \end{axis}

    \begin{axis}[
        name=simulationKlambdaDIANA,
        at={(simulationPlaneDIANA.south east)},
        anchor=south west,
        xshift = 2cm,
        height=\heightAll,
        width = \widthKq,
        xmajorgrids, 
        ymajorgrids,
        xlabel={$t$ (s)},
        xmin=0,
        xmax=10,
        ylabel={overlap metric $\kappa^\star_i$}, 
        legend cell align={left},
        legend columns = 1,
        legend style={
            at={(1.0,0.0)},
            anchor=south east,
            },
        ]
 
        \addplot[color = Kqplot, line width = 0.4mm, each nth point=1, line join=bevel] table[x index=0,y index=1]{src/data/SimExampleDIANA/Kq_20241120_173405.txt}; \addlegendentry{$\kappa^{\star}_{1}$}
        \addplot[color = teal, line width = 0.4mm, each nth point=1, line join=bevel] table[x index=0,y index=2]{src/data/SimExampleDIANA/Kq_20241120_173405.txt}; \addlegendentry{$\kappa^{\star}_{2}$}
        
    \end{axis}

    \begin{axis}[
        name=simulationComptimeDIANA,
        at={(simulationKlambdaDIANA.south east)},
        anchor=south west,
        xshift = 1.75cm,
        height=\heightAll,
        width = \widthT,
        xmajorgrids, 
        ymajorgrids,
        xlabel={$t$ (s)},
        ylabel={solution time (ms)},
        ymode=log,
        ymax = 500,
        log ticks with fixed point, 
        ymin = 1,
        xmin=0,
        xmax=10,
        ]
 
        \addplot[color = blue, only marks, mark = x, line width = 0.2mm, each nth point=1] table[x index=0, y expr=1e3*\thisrowno{1}]{src/data/SimExampleDIANA/CompTime_20241120_173405.txt};
        \addplot[domain=0:20, samples=10, color=gray, thick] {200}; 
        \node[gray] at (7.25,290) {$\delta_t$};

    \end{axis}
\end{tikzpicture}

%% file: figures/lab_png.tex
\begin{tikzpicture}
\node[inner sep=0pt] (sevenPlot) at (0,0)
    {\includegraphics[height = 3cm, trim = 100 100 150 100, clip]{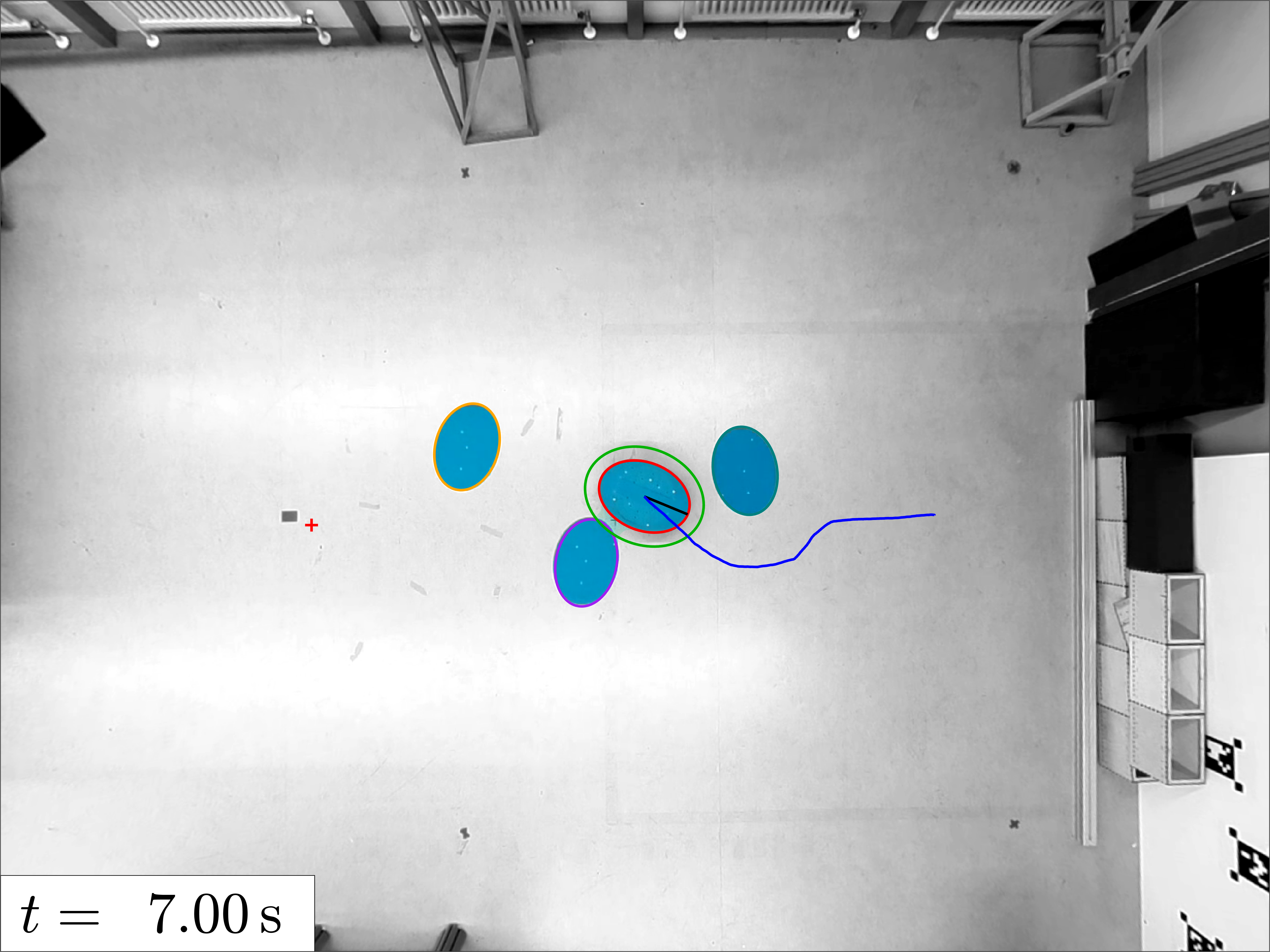}};
\node[inner sep=0pt, outer sep=2pt, anchor = south west] (sevenTime) at (sevenPlot.south west) {$t=\unit[7]{s}$};

\node[inner sep=0pt, anchor=south west, xshift = 0.5cm] (elevenPlot) at (sevenPlot.south east)
{\includegraphics[height = 3cm, trim = 100 100 150 100, clip]{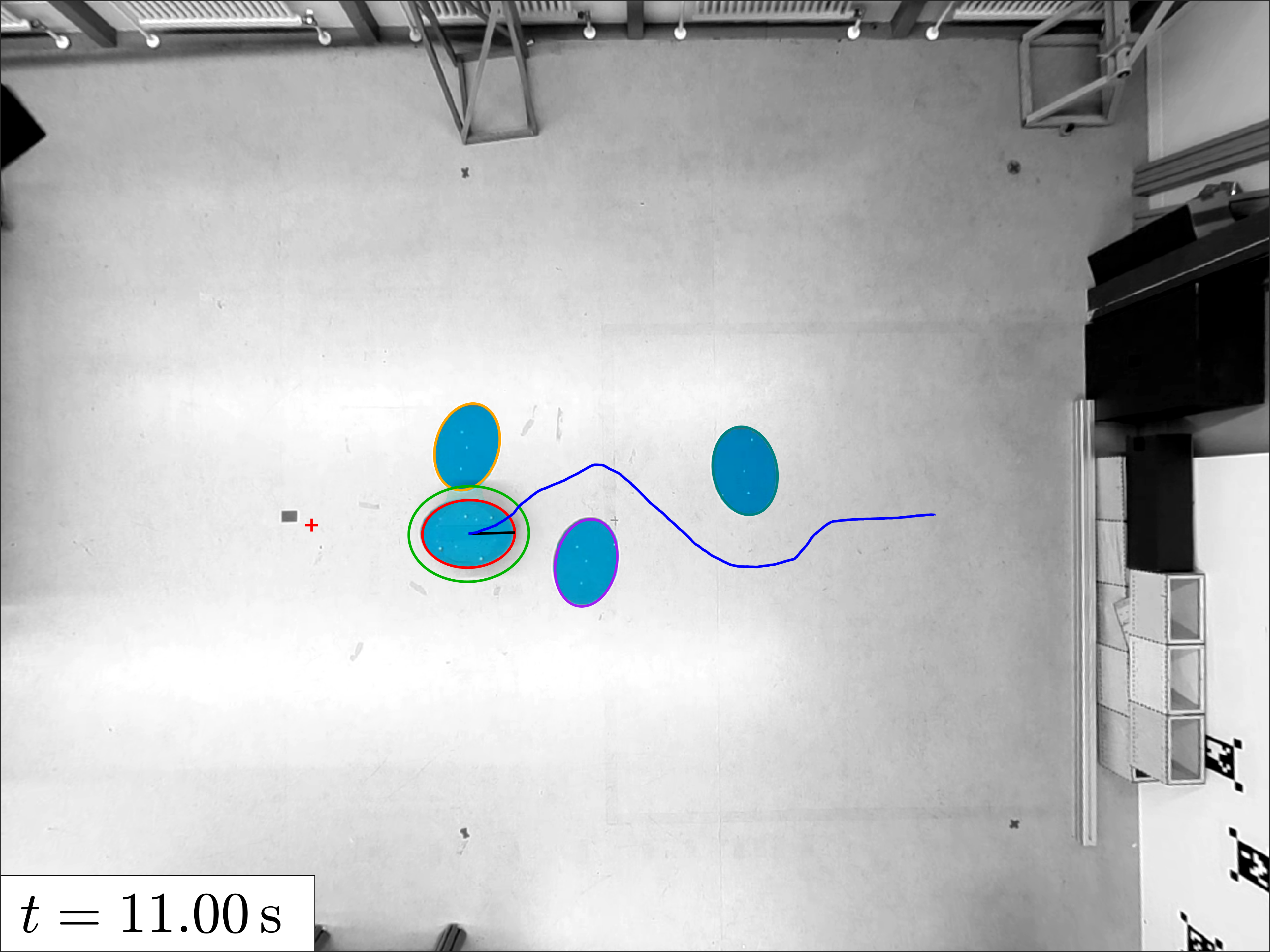}};
\node[inner sep=0pt, outer sep=2pt, anchor = south west] (elevenTime) at (elevenPlot.south west) {$t=\unit[11]{s}$};

\begin{axis}[
        name=KqHardwareExperiment,
        at={(elevenPlot.north east)},
        anchor=north west,
        xshift = 2.4cm,
        height= 3.75cm,
        width = 5cm,
        xmajorgrids, 
        ymajorgrids,
        xlabel={$t$ in (s)},
        ylabel style={align=center}, 
        ylabel={overlap \\ metric $\kappa^\star_i$}, 
        legend cell align={left},
        legend columns = 1,
        legend style={
            at={(1.075,1.0)},
            anchor=north west,
            },
        ymin = -800,
        ymax = 100,
        ]
        \addplot[color = Kqplot, line width = 0.4mm, each nth point=1, line join=bevel] table[x index=0,y index=1]{src/data/20240306_172538_Kq_post.txt}; \addlegendentry{$\kappa^{\star}_{1,\textnormal{o}}$}
        \addplot[dashed, color = Kqplot, line width = 0.4mm, each nth point=1, line join=bevel] table[x index=0,y index=1]{src/data/20240306_172538_Kq_post_nosafe.txt}; \addlegendentry{$\kappa^{\star}_{1,\textnormal{r}}$}
        \addplot[color = teal, line width = 0.4mm, each nth point=1, line join=bevel] table[x index=0,y index=2]{src/data/20240306_172538_Kq_post.txt}; \addlegendentry{$\kappa^{\star}_{2,\textnormal{o}}$}
        \addplot[dashed, color = teal, line width = 0.4mm, each nth point=1, line join=bevel] table[x index=0,y index=2]{src/data/20240306_172538_Kq_post_nosafe.txt}; \addlegendentry{$\kappa^{\star}_{2,\textnormal{r}}$}
        \addplot[color = orange, line width = 0.4mm, each nth point=1, line join=bevel] table[x index=0,y index=3]{src/data/20240306_172538_Kq_post.txt}; \addlegendentry{$\kappa^{\star}_{3,\textnormal{o}}$}
        \addplot[dashed, color = orange, line width = 0.4mm, each nth point=1, line join=bevel] table[x index=0,y index=3]{src/data/20240306_172538_Kq_post_nosafe.txt}; \addlegendentry{$\kappa^{\star}_{3,\textnormal{r}}$}
    \end{axis}

\end{tikzpicture}